\pdfoutput=1

\documentclass[11pt]{article}

\usepackage[]{emnlp2021}

\usepackage{times}
\usepackage{latexsym}
\usepackage{amsfonts,amsmath,amssymb,bm}
\usepackage{times}
\usepackage{multirow}
\usepackage{graphicx}

\usepackage{booktabs}
\usepackage{xcolor}
\usepackage{multirow}
\usepackage{setspace}
\usepackage{pdflscape}
\usepackage{enumitem}
\usepackage{rotating}

\usepackage{array}
\newcommand{\PreserveBackslash}[1]{\let\temp=\\#1\let\\=\temp}
\newcolumntype{C}[1]{>{\PreserveBackslash\centering}p{#1}}

\usepackage[T1]{fontenc}

\usepackage[utf8]{inputenc}

\usepackage{microtype}

\title{Distilling Relation Embeddings from Pre-trained Language Models}

\author{Asahi Ushio \and Jose Camacho-Collados \and Steven Schockaert\\
  Cardiff NLP, School of Computer Science and Informatics \\ Cardiff University, United Kingdom\\
  \texttt{\{UshioA,CamachoColladosJ,SchockaertS1\}@cardiff.ac.uk}
  \\}

\begin{document}
\maketitle

\begin{abstract}
Pre-trained language models have been found to capture a surprisingly rich amount of lexical knowledge, ranging from commonsense properties of everyday concepts to detailed factual knowledge about named entities. Among others, this makes it possible to distill high-quality word vectors from pre-trained language models. However, it is currently unclear to what extent it is possible to distill \emph{relation embeddings}, i.e.\ vectors that characterize the relationship between two words. Such relation embeddings are appealing because they can, in principle, encode relational knowledge in a more fine-grained way than is possible with knowledge graphs. To obtain relation embeddings from a pre-trained language model, we encode word pairs using a (manually or automatically generated) prompt, and we fine-tune the language model such that relationally similar word pairs yield similar output vectors. We find that the resulting relation embeddings are highly competitive on analogy (unsupervised) and relation classification (supervised) benchmarks, even without any task-specific fine-tuning.\footnote{Source code to reproduce our experimental results and the model checkpoints are available in the following repository:  \url{https://github.com/asahi417/relbert}}
\end{abstract}

\section{Introduction}
One of the most widely studied aspects of word embeddings is the fact that word vector differences capture lexical relations \cite{mikolov2013distributed}. 
While not being directly connected to downstream performance on NLP tasks, this ability of word embeddings is nonetheless important. For instance, understanding lexical relations is an important prerequisite for understanding the meaning of compound nouns \cite{turney2012domain}. Moreover, the ability of word vectors to capture semantic relations has enabled a wide range of applications beyond NLP, including flexible querying of relational databases \cite{bordawekar2017using}, schema matching \cite{fernandez2018seeping}, completion and retrieval of Web tables \cite{zhang2019table2vec}, ontology completion \cite{DBLP:conf/aaai/BouraouiS19} and information retrieval in the medical domain \cite{info:doi/10.2196/16948}. More generally, relational similarity (or analogy) plays a central role in computational creativity \cite{goel2019computational}, legal reasoning \cite{ashley1988arguing,walton2010similarity}, ontology alignment \cite{raad2015role} and instance-based learning \cite{miclet2008analogical}.

Given the recent success of pre-trained language models \cite{devlin-etal-2019-bert,RoBERTa,GPT3}, we may wonder whether such models are able to capture lexical relations in a more faithful or fine-grained way than traditional word embeddings. However, for language models (LMs), there is no direct equivalent to the word vector difference. In this paper, we therefore propose a strategy for extracting relation embeddings from pre-trained LMs, i.e.\ vectors encoding the relationship between two words. On the one hand, this will allow us to gain a better understanding of how well lexical relations are captured by these models. On the other hand, this will also provide us with a practical method for obtaining relation embeddings in applications such as the ones mentioned above.

Since it is unclear how LMs store relational knowledge, rather than directly extracting relation embeddings, we first fine-tune the LM, such that relation embeddings can be obtained from its output. To this end, we need a prompt, i.e.\ a template to convert a given word pair into a sentence, and some training data to fine-tune the model. To illustrate the process, consider the word pair \emph{Paris}-\emph{France}. As a possible input to the model, we could use a sentence such as ``The relation between Paris and France is <mask>". Note that our aim is to find a strategy that can be applied to any pair of words, hence the way in which the input is represented needs to be sufficiently generic. We then fine-tune the LM such that its output corresponds to a relation embedding. To this end, we use a crowdsourced dataset of relational similarity judgements that was collected in the context of SemEval 2012 Task 2 \cite{jurgens-etal-2012-semeval}. Despite the relatively small size of this dataset, we show that the resulting fine-tuned LM allows us to produce high-quality relation embeddings, as confirmed in our extensive evaluation in analogy and relation classification tasks. Importantly, this also holds for relations that are of a different nature than those in the SemEval dataset, showing that this process allows us to distill relational knowledge that is encoded in the pre-trained LM, rather than merely generalising from the examples that were used for fine-tuning.

\section{Related Work}
\paragraph{Probing LMs for Relational Knowledge}
Since the introduction of transformer-based LMs, a large number of works have focused on analysing the capabilities of such models, covering the extent to which they capture syntax \cite{goldberg2019assessing,saphra-lopez:2019,hewitt-manning:2019,van-schijndel-etal-2019-quantity,jawahar-etal-2019-bert,Tenney-et-al-2019}, lexical semantics \cite{DBLP:conf/emnlp/Ethayarajh19,DBLP:conf/acl/BommasaniDC20,DBLP:conf/emnlp/VulicPLGK20}, and various forms of factual and commonsense knowledge \cite{petroni-etal-2019-language,DBLP:conf/cogsci/ForbesHC19,davison2019commonsense,zhou2020evaluating,talmor2020olmpics,roberts2020much}, among others. The idea of extracting relational knowledge from LMs, in particular, has also been studied. For instance, \citet{petroni-etal-2019-language} use BERT for link prediction. To this end, they use a manually defined prompt for each relation type, in which the tail entity is replaced by a <mask> token. To complete a knowledge graph triple such as (\emph{Dante}, \emph{born-in}, ?) they create the input ``\emph{Dante was born in <mask>}'' and then look at the predictions of BERT for the masked token to retrieve the correct answer. It is notable that BERT is thus used for extracting relational knowledge without any fine-tuning. This clearly shows that a substantial amount of factual knowledge is encoded in the parameters of pre-trained LMs. Some works have also looked at how such knowledge is stored. \citet{geva2020transformer} argue that the feed-forward layers of transformer-based LMs act as neural memories, which would suggest that e.g.\ ``the place where Dante is born'' is stored as a property of Florence. \citet{dai2021knowledge} present further evidence of this view. What is less clear, then, is whether relations themselves have an explicit representation, or whether transformer models essentially store a propositionalised knowledge graph. The results we present in this paper suggest that common lexical relations (e.g.\ hypernymy, meronymy, has-attribute), at least, must have some kind of explicit representation, although it remains unclear how they are encoded.

Another notable work focusing on link prediction is \cite{bosselut2019comet}, where GPT is fine-tuned to complete triples from commonsense knowledge graphs, in particular ConceptNet \cite{conceptnet2017} and ATOMIC \cite{sap2019atomic}. While their model was able to generate new knowledge graph triples, it is unclear to what extent this is achieved by extracting commonsense knowledge that was already captured by the pre-trained GPT model, or whether this rather comes from the ability to generalise from the training triples. For the ConceptNet dataset, for instance, \citet{jastrzkebski2018commonsense} found that most test triples are in fact minor variations of training triples. In this paper, we also rely on fine-tuning, which makes it harder to determine to what extent the pre-trained LM already captures relational knowledge. We address this concern by including relation types in our evaluation which are different from the ones that have been used for fine-tuning.

\paragraph{Unsupervised Relation Discovery}
Modelling how different words are related is a long-standing challenge in NLP. An early approach is DIRT \cite{Lin2001}, which encodes the relation between two nouns as the dependency path connecting them. Their view is that two such dependency paths are similar if the sets of word pairs with which they co-occur are similar. \citet{DBLP:conf/acl/HasegawaSG04} cluster named entity pairs based on the bag-of-words representations of the contexts in which they appear. Along the same lines, \newcite{Yao2011} proposed a generative probabilistic model, inspired by LDA \cite{DBLP:journals/jmlr/BleiNJ03}, in which relations are viewed as latent variables (similar to topics in LDA). \citet{Turney:2005:MSS:1642293.1642475} proposed a method called Latent Relational Analysis (LRA), which uses matrix factorization to learn relation embeddings based on co-occurrences of word pairs and dependency paths. Matrix factorization is also used in the Universal Schema approach from Riedel et al.\ \cite{Riedel2013}, which jointly models the contexts in which words appear in a corpus with a given set of relational facts.

\setlength{\parskip}{0.4em} 

The aforementioned works essentially represent the relation between two words by summarising the contexts in which these words co-occur. In recent years, a number of strategies based on distributional models have been explored that rely on similar intuitions but go beyond simple vector operations of word embeddings.\footnote{Interestingly, \newcite{roller-erk-2016-relations} showed that the direct concatenation of distributional word vectors in isolation can effectively identify Hearst Patterns \cite{hearst-1992-automatic}.} For instance, \citet{Jameel2018} introduced a variant of the GloVe word embedding model, in which relation vectors are jointly learned with word vectors. In SeVeN \cite{espinosa-anke-schockaert-2018-seven} and RELATIVE \cite{camachocollados:ijcai2019relative}, relation vectors are computed by averaging the embeddings of context words, while pair2vec \cite{joshi-etal-2019-pair2vec} uses an LSTM to summarise the contexts in which two given words occur, and \citet{Washio2018a} learn embeddings of dependency paths to encode word pairs. 
Another line of work is based on the idea that relation embeddings should facilitate link prediction, i.e.\ given the first word and a relation vector, we should be able to predict the second word \cite{Marcheggiani2016,DBLP:conf/acl/SimonGP19}. This idea also lies at the basis of the approach from \citet{Soares2019}, who train a relation encoder by fine-tuning BERT \cite{devlin-etal-2019-bert} with a link prediction loss. However, it should be noted that they focus on learning relation vectors from individual sentences, as a pre-training task for applications such as few-shot relation extraction. In contrast, our focus in this paper is on characterising the overall relationship between two words. 

\section{RelBERT}

\begin{figure}[t]
    \centering
    \includegraphics[width=\columnwidth]{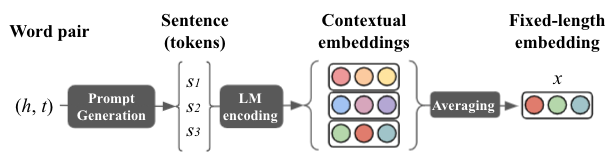}
    \caption{Pipeline to transform the word pair $(h,t)$ to the relation embedding $\bm{x}$.}
    \label{fig:image_embedding}
\end{figure}

In this section, we describe our proposed relation embedding model (\textit{RelBERT} henceforth). 
To obtain a relation embedding for given a word pair $(h, t)$, we first convert it into a sentence $s$, called the prompt. We then feed the prompt through the LM and average the contextualized embeddings (i.e.\ the output vectors) to get the relation embedding of $(h,t)$. These steps are illustrated in Figure~\ref{fig:image_embedding} and explained in more detail in the following.

\subsection{Prompt Generation}\label{sec:prompt-generation}
\paragraph{Manual Prompts}
A basic prompt generation strategy is to rely on manually created templates, which has proven effective in factual knowledge probing \cite{petroni-etal-2019-language} and text classification \cite{schick-schutze-2021-exploiting,tam2021improving,le2021many}, among many others.
To test whether manually generated templates can be effective for learning relation embeddings, we will consider the following five templates:

{\small
\begin{enumerate}[topsep=0pt,itemsep=0ex,partopsep=1ex,parsep=1ex]
    {
    \item  Today, I finally discovered the relation between \textbf{[h]} and \textbf{[t]} : \textbf{[h]} is the <mask> of \textbf{[t]}
    \item Today, I finally discovered the relation between \textbf{[h]} and \textbf{[t]} : \textbf{[t]} is \textbf{[h]}'s <mask>
    \item Today, I finally discovered the relation between \textbf{[h]} and \textbf{[t]} : <mask>
    \item I wasn’t aware of this relationship, but I just read in the encyclopedia that \textbf{[h]} is the <mask> of \textbf{[t]}
    \item I wasn’t aware of this relationship, but I just read in the encyclopedia that \textbf{[t]} is \textbf{[h]}’s <mask>
  }
\end{enumerate}
}

\noindent where {\it <mask>} is the LM's mask token, and \textbf{[h]} and \textbf{[t]} are slots that are filled with the head word $h$ and tail word $t$ from the given word pair $(h,t)$ respectively. The main intuition is that the template should encode that we are interested in the relationship between $h$ and $t$. Moreover, we avoid minimal templates such as ``\textbf{[h]} is the {\it <mask>} of \textbf{[t]}'', as LMs typically perform worse on such short inputs \cite{bouraoui2020inducing,jiang-etal-2020-know}.

\paragraph{Learned Prompts}
The choice of prompt can have a significant impact on an LM's performance. Since it is difficult to generate manual prompts in a systematic way, several strategies for automated generation of task-specific prompts have been proposed, e.g.\ based on
mining patterns from a corpus \cite{bouraoui2020inducing}, paraphrasing \cite{jiang-etal-2020-know}, training an additional LM for template generation \cite{haviv-etal-2021-bertese,gao2020making},
and prompt optimization \cite{shin-etal-2020-autoprompt,liu2021gpt}.
In our work, we focus on the latter strategy, given its conceptual simplicity and its strong reported performance on various benchmarks. Specifically, we consider AutoPrompt \cite{shin-etal-2020-autoprompt} and P-tuning \cite{liu2021gpt}. Note that both methods rely on training data. We will use the same training data and loss function that we use for fine-tuning the LM; see Section \ref{sec:finetuning}.

AutoPrompt initializes the prompt as a fixed-length template: 
\begin{align}
    T = (&z_1, \dots, z_{\pi}, \textbf{[h]}, z_{\pi +1}, \dots, z_{\pi + \tau}, \nonumber\\
    &\textbf{[t]}, z_{\pi + \tau + 1}, \dots, z_{\pi + \tau + \gamma} )
\end{align}
where $\pi$, $\tau$, $\gamma$ are hyper-parameters which determine the length of the template. The tokens of the form $z_i$ are called trigger tokens. These tokens are initialized as {\it <mask>}. The method then iteratively finds the best token to replace each mask, based on the gradient of the task-specific loss function.\footnote{We note that in most implementations of AutoPrompt the vocabulary to sample trigger tokens is restricted to that of the training data. However, given the nature of our training data (i.e., pairs of words and not sentences), we consider the full pre-trained LM's vocabulary.}

P-tuning employs the same template initialization as AutoPrompt but its trigger tokens are newly introduced special tokens with trainable embeddings $\bm{\hat{e}}_{1:\pi + \tau + \gamma}$, which are learned using a task-specific loss function while the LM's weights are frozen.

\subsection{Fine-tuning the LM}\label{sec:finetuning}
To fine-tune the LM, we need training data and a loss function. As training data, we assume that, for a number of different relation types $r$, we have access to examples of word pairs $(h,t)$ that are instances of that relation type.
The loss function is based on the following intuition: the embeddings of word pairs that belong to the same relation type should be closer together than the embeddings of pairs that belong to different relations. In particular, we use the triplet loss from \citet{schroff2015facenet} and the classification loss from \newcite{reimers-gurevych-2019-sentence}, both of which are based on this intuition. 

\paragraph{Triplet Loss}
We draw a triplet from the relation dataset by selecting an anchor pair $a=(h_a,t_a)$, a positive example $p=(h_p,t_p)$ and a negative example $n=(h_n,t_n)$, i.e.\ we select word pairs $a,p,n$ such that $a$ and $p$ belong to the same relation type while $n$ belongs to a different relation type. Let us write $\bm{x}_a$, $\bm{x}_p$, $\bm{x}_n$ for the corresponding relation embeddings.
Each relation embedding is produced by the same LM, which is trained to make the distance between $\bm{x}_a$ and $\bm{x}_p$ smaller than the distance between $\bm{x}_a$ and $\bm{x}_n$. Formally, this is accomplished using the following triplet loss function:
\begin{align*}
    L_t = \max\big(0, \|\bm{x}_a-\bm{x}_p\| - \|\bm{x}_a - \bm{x}_n\| + \varepsilon \big)
\end{align*}
where $\varepsilon>0$ is the margin and $\|\cdot\|$ is the $l^2$ norm.

\paragraph{Classification Loss} Following SBERT \cite{reimers-gurevych-2019-sentence}, we use a classifier to predict whether two word pairs belong to the same relation. 
The classifier is jointly trained with the LM using the negative log likelihood loss function:
\begin{align*}
    L_{c} = -\log(g(\bm{x}_a, \bm{x}_p)) - \log(1-g(\bm{x}_a, \bm{x}_n))
\end{align*}
where
\begin{align*}
    g(\bm{u}, \bm{v}) = \text{sigmoid}(W \cdot \left[ \bm{u} \oplus \bm{v} \oplus |\bm{v} - \bm{u}| \right]^{T} )
\end{align*}
with $W\in \mathbb{R}^{3\times d}$, $\bm{u},\bm{v}\in \mathbb{R}^d$, $|\cdot|$ the element-wise absolute difference, and $\oplus$ concatenation.

\section{Experimental Setting}

In this section, we explain our experimental setting to train and evaluate RelBERT. 

\subsection{RelBERT Training}\label{sec:relbert-training}
\paragraph{Dataset}
We use the platinum ratings from SemEval 2012 Task 2 \cite{jurgens-etal-2012-semeval} as our training dataset for RelBERT. This dataset covers 79 fine-grained semantic relations, which are grouped in 10 categories. For each of the 79 relations, the dataset contains a typicality score for a number of word pairs (around 40 on average),
indicating to what extent the word pair is a prototypical instance of the relation. We treat the top 10 pairs (i.e.\ those with the highest typicality score) as positive examples of the relation, and the bottom 10 pairs as negative examples. We use 80\% of these positive and negative examples for training RelBERT (i.e.\ learning the prompt and fine-tuning the LM) and 20\% for validation.

\begin{figure}[t]
    \centering
    \includegraphics[width=\columnwidth]{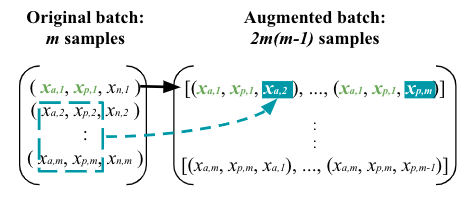}
    \caption{Batch augmentation where the original batch with $m$ samples is augmented with $2m(m-1)$ samples.}
    \label{fig:image_batch}
\end{figure}

\paragraph{Constructing Training Triples}
We rely on three different strategies for constructing training triples. First, we obtain triples by selecting two positive examples of a given relation type (i.e.\ from the top-10 pairs) and one negative example (i.e.\ from the bottom-10 pairs). We construct 450 such triples per relation. Second, we construct triples by using two positive examples of the relation and one positive example from another relation (which is assumed to correspond to a negative example). In particular, for efficiency, we use the anchors and positive examples of the other triples from the same batch as negative examples (while ensuring that these triples are from different relations). Figure~\ref{fig:image_batch} illustrates this idea. Note how the effective batch size thus increases quadratically, while the number of vectors that needs to be encoded by the LM remains unchanged. In our setting, this leads to an additional
13500 triples per relation.
Similar in-batch negative sampling has been shown to be effective in information retrieval  \cite{karpukhin-etal-2020-dense,gillick-etal-2019-learning}. Third, we also construct training triples by considering the 10 high-level categories as relation types. In this case, we choose two positive examples from different relations that belong to the same category, along with a positive  example from a relation from a different category.
We add 5040 triples of this kind for each of the 10 categories.

\paragraph{Training} RelBERT training consists of two phases: prompt optimization (unless a manually defined prompt is used) and language model fine-tuning. First we optimize the prompt over the training set with the triplet loss $L_t$ while the parameters of the LM are frozen. Subsequently, we fine-tune the LM with the resulting prompt, using the sum of the triplet loss $L_t$ and the classification loss $L_c$ over the same training set. We do not use the classification loss during the prompt optimisation, as that would involve training the classifier while optimizing the prompt. 
We select the best hyper-parameters of the prompting methods based on the final loss over the validation set. In particular, when manual prompts are used, we choose the best template among the five candidates described in Section~\ref{sec:prompt-generation}. For AutoPrompt and P-tuning, we consider all combinations of $\pi\in\{8,9\}$, $\tau\in\{1, 2\}$, $\gamma\in\{1, 2\}$.
We use RoBERTa \cite{RoBERTa} as our main LM, where the initial weights were taken from the \texttt{roberta-large} model checkpoint shared by the Huggingface transformers model hub \citep{wolf-etal-2020-transformers}.
We use the Adam optimizer \cite{kingma2014adam} with learning rate 0.00002, batch size 64 and we fine-tune the model for 1 epoch. For AutoPrompt, the top-50 tokens are considered and the number of iterations is set to 50. In each iteration, one of the input tokens is re-sampled and the loss is re-computed across the entire training set.\footnote{We should note that AutoPrompt takes considerably longer than any other components of RelBERT. More details on experimental training times are included in the appendix.} For P-tuning, we train the weights that define the trigger embeddings (i.e.\ the weights of the input vectors and the parameters of the LSTM) for 2 epochs.
Note that we do not tune RelBERT on any task-specific training or validation set. We thus use the same relation embeddings across all the considered evaluation tasks.

\subsection{Evaluation Tasks}
We evaluate RelBERT on two relation-centric tasks: solving analogy questions (unsupervised) and  lexical relation classification (supervised). 

\paragraph{Analogy Questions}
We consider the task of solving word analogy questions. Given a query word pair, the model is required to select the relationally most similar word pair from a list of candidates. To solve this task, we simply choose the candidate whose RelBERT embedding has the highest cosine similarity with the RelBERT embedding of the query pair. Note that this task is completely unsupervised, without the need for any training or tuning. 
We use the five analogy datasets that were considered by \citet{ushio-etal-2021-bert-is}: the SAT analogies dataset \cite{DBLP:conf/ranlp/TurneyLBS03}, the U2 and U4 analogy datasets, which were collected from an educational website\footnote{\url{https://englishforeveryone.org/Topics/Analogies.html}}, and datasets that were derived\footnote{In particular, they were converted into the same format of multiple-choice questions as the other datasets.} from
BATS \cite{gladkova-etal-2016-analogy} and the Google analogy dataset \cite{mikolov-etal-2013-linguistic}. These five datasets consist of tuning and testing fragments. 
In particular, they contain 37/337 (SAT), 24/228 (U2), 48/432 (U4), 50/500 (Google), and 199/1799 (BATS) questions for validation/testing.
As there is no need to tune RelBERT on task-specific data, we only use the test fragments.
For SAT, we will also report results on the full dataset (i.e.\ the testing fragment and tuning fragment combined), as this allows us to compare the performance with published results. We will refer to this full version of the SAT dataset as SAT\dag.

\begin{table*}[t]
\centering
\scalebox{0.85}{
\begin{tabular}{lrrrrr}
\toprule
\textbf{}   & \multicolumn{1}{c}{\textbf{BLESS}} & \multicolumn{1}{c}{\textbf{CogALex}} & \multicolumn{1}{c}{\textbf{EVALution}} & \multicolumn{1}{c}{\textbf{K\&H+N}} & \multicolumn{1}{c}{\textbf{ROOT09}} \\ \midrule
Random            &  8,529/609/3,008 &  2,228/3,059 &             - &  18,319/1,313/6,746 &  4,479/327/1,566 \\
Meronym           &    2,051/146/746 &      163/224 &     218/13/86 &          755/48/240 &                - \\
Event             &    2,657/212/955 &            - &             - &                   - &                - \\
Hypernym          &       924/63/350 &      255/382 &  1,327/94/459 &     3,048/202/1,042 &    2,232/149/809 \\
Co-hyponym       &    2,529/154/882 &            - &             - &  18,134/1,313/6,349 &    2,222/162/816 \\
Attribute         &    1,892/143/696 &            - &    903/72/322 &                   - &                - \\
Possession &                - &            - &    377/25/142 &                   - &                - \\
Antonym           &                - &      241/360 &  1,095/90/415 &                   - &                - \\
Synonym           &                - &      167/235 &    759/50/277 &                   - &                - \\
\bottomrule
\end{tabular}
}
\caption{Number of instances for each relation type across training/validation/test sets of all lexical relation classification datasets.
\label{tab:data_stats}}
\end{table*}

\paragraph{Lexical Relation Classification} We consider the task of predicting which relation a given word pair belongs to. To solve this task, we train a multi-layer perceptron (MLP) which takes the (frozen) RelBERT embedding of the word pair as input.
We consider the following widely-used multi-class relation classification benchmarks:
K\&H+N \cite{necsulescu-etal-2015-reading},
BLESS \cite{baroni-lenci-2011-blessed},
ROOT09 \cite{santus-etal-2016-nine},
EVALution \cite{santus-etal-2015-evalution}, and
 CogALex-V Subtask 2 \cite{santus-etal-2016-cogalex}.
 Table~\ref{tab:data_stats} shows the size of the training, validation and test  sets for each of the relation classification dataset. The hyperparameters of the MLP classifier are tuned on the validation set of each dataset. Concretely, we tune the learning rate from $[0.001, 0.0001, 0.00001]$ and the hidden layer size from $[100, 150, 200]$.
CogALex-V only has testing fragments so for this dataset we employ the default configuration of Scikit-Learn \cite{JMLR:v12:pedregosa11a}, which uses a 100-dimensional hidden layer and is optimized using Adam with a learning rate of 0.001.
These datasets focus on the following lexical relations: co-hyponymy (cohyp),  hypernymy (hyp), meronymy (mero), possession (poss), synonymy (syn),  antonymy (ant), attribute (attr), event, and random (rand).

\subsection{Baselines}
As baselines, we consider two standard word embedding models: GloVe \cite{pennington-etal-2014-glove} and FastText \cite{bojanowski-etal-2017-enriching}, where word pairs are represented by the vector difference of their word embeddings (\textit{diff}).\footnote{Vector difference is the most common method for encoding relations, and has been shown to be the most reliable in the context of word analogies \cite{hakami2017compositional}.}
For the classification experiments, we also consider the concatenation of the two word embeddings (\textit{cat}) and their element-wise multiplication\footnote{Multiplicative features have been shown to provide consistent improvements for word embeddings in supervised relation classification tasks \cite{vu-shwartz-2018-integrating}.} (\textit{dot}). We furthermore experiment with two pre-trained word pair embedding models: pair2vec \cite{joshi-etal-2019-pair2vec} (\textit{pair}) and RELATIVE \cite{camachocollados:ijcai2019relative} (\textit{rel}).
For these word pair embeddings, as well as for RelBERT, we concatenate the embeddings from both directions, i.e.\ $(h, t)$ and $(t, h)$.
For the analogy questions, two simple statistical baselines are included: the expected random performance and a strategy based on point-wise mutual information (PMI) \citet{church1990word}. In particular, the PMI score of a word pair is computed using the English Wikipedia, with a fixed window size of 10. We then choose the candidate pair with the highest PMI as the prediction. Note that this PMI-based method completely ignores the query pair.
We also compare with the published results from \citet{ushio-etal-2021-bert-is}, where a strategy is proposed to solve analogy questions by using LMs to compute an \emph{analogical proportion score}. In particular,  a four-word tuple  $(a,b,c,d)$ is encoded using a custom prompt and perplexity based scoring strategies are used to determine whether the word pair $(a,b)$ has the same relation as the word pair $(c,d)$.
Finally, for the {SAT\dag} dataset, we compare with the published results from GPT-3 \cite{GPT3}, LRA \cite{Turney:2005:MSS:1642293.1642475} and SuperSim \cite{turney-2013-distributional}; for relation classification we report the published results of the LexNet \cite{shwartz-etal-2016-improving} and SphereRE \cite{wang-etal-2019-spherere} relation classification models, taking the results from the latter publication. We 
did not reproduce these latter methods in similar conditions as our work, and hence they are not fully comparable. Moreover, these approaches are a different nature, as the aim of our work is to provide universal relation embeddings instead of task-specific models.

\section{Results}
In this section, we present our main experimental results, testing the relation embeddings learned by RelBERT on analogy questions (Section \ref{eval-analogy}) and relation classification (Section \ref{eval-relclass}).

\subsection{Analogy Questions}
\label{eval-analogy}

Table~\ref{tab:analogy_result} shows the accuracy on the analogy benchmarks. The RelBERT models substantially outperform the baselines on all datasets, except for the Google analogy dataset.\footnote{The Google analogy dataset has been shown to be biased toward word similarity and therefore to be well suited to word embeddings \cite{linzen-2016-issues,rogers-etal-2017-many}.} 
Comparing the different prompt generation approaches, we can see that, surprisingly, the manual prompt consistently outperforms the automatically-learned prompt strategies.

On SAT\dag, RelBERT outperforms LRA, which represents the state-of-the-art in the zero-shot setting, i.e.\ in the setting where no training data from the SAT dataset is used. RelBERT moreover outperforms GPT-3 in the few-shot setting, despite not using any training examples. In contrast, GPT-3 encodes a number of training examples as part of the prompt.

It can furthermore be noted that the other two relation embedding methods (i.e.\ pair2vec and RELATIVE) perform poorly in this unsupervised task. The analogical proportion score from \citet{ushio-etal-2021-bert-is} also underperforms RelBERT, even when tuned on dataset-specific tuning data.

\begin{table}[t]
\centering
\resizebox{\columnwidth}{!}{
\begin{tabular}{l@{\hspace{5pt}}C{3em}@{}C{3em}@{}C{3em}@{}C{3em}@{}C{3em}@{}C{3em}}
\toprule
\textbf{Model}      & \textbf{SAT\dag} & \textbf{SAT}  & \textbf{U2}   & \textbf{U4}   & \textbf{Google} & \textbf{BATS} \\\midrule
Random              & 20.0              & 20.0          & 23.6          & 24.2          & 25.0            & 25.0          \\
PMI                 & 23.3              & 23.1          & 32.9          & 39.1          & 57.4            & 42.7          \\
LRA                 & \textit{56.4}              & -             & -             & -             & -               & -             \\
SuperSim        & \textit{54.8}              & -             & -             & -             & -               & -             \\
GPT-3 (zero)    & \textit{53.7}              & -             & -             & -             & -               & -             \\
GPT-3 (few)     & \textit{65.2}*              & -             & -             & -             & -               & -             \\
RELATIVE           &              24.9 &          24.6 &         32.5 &         27.1 &             62.0 &           39.0 \\
pair2vec           &              33.7 &          34.1 &         25.4 &         28.2 &             66.6 &           53.8 \\
GloVe              &              48.9 &          47.8 &         46.5 &         39.8 &             96.0 &           68.7 \\
FastText           &              49.7 &          47.8 &         43.0 &         40.7 &             \textbf{\underline{96.6}} &           72.0 \\\midrule
\multicolumn{7}{l}{Analogical Proportion Score}\\
$\cdot$ GPT-2       & \textit{41.4}    & \textit{35.9}& \textit{41.2}    & \textit{44.9}    & \textit{80.4}& \textit{63.5} \\
$\cdot$ BERT        & \textit{32.6}    & \textit{32.9}& \textit{32.9}    & \textit{34.0}    & \textit{80.8}& \textit{61.5} \\
$\cdot$ RoBERTa     & \textit{49.6}    & \textit{42.4}& \textit{49.1}    & \textit{49.1}    & \textit{90.8}& \textit{69.7} \\\midrule
\multicolumn{7}{l}{Analogical Proportion Score (tuned)}\\
$\cdot$ GPT-2       & \textit{57.8}*    & \textit{56.7}*& \textit{50.9}*    & \textit{49.5}*    & \textit{95.2}*& \textit{\underline{81.2}}* \\
$\cdot$ BERT       & \textit{42.8}*    & \textit{41.8}*& \textit{44.7}*    & \textit{41.2}*    & \textit{88.8}*& \textit{67.9}* \\
$\cdot$ RoBERTa     & \textit{55.8}*    & \textit{53.4}*& \textit{58.3}*    & \textit{57.4}*    & \textit{93.6}*& \textit{78.4}* \\\midrule
\multicolumn{7}{l}{RelBERT} \\
$\cdot$ Manual     &              \textbf{\underline{69.5}} &\textbf{\underline{70.6}} & \textbf{\underline{66.2}} &\textbf{\underline{65.3}} &             92.4 &           \textbf{78.8} \\
$\cdot$ AutoPrompt &              61.0 &          62.3 &         61.4 &         63.0 &             88.2 &           74.6 \\
$\cdot$ P-tuning   &              54.0 &          55.5 &         58.3 &         55.8 &             83.4 &           72.1 \\\bottomrule
\end{tabular}
}
\caption{Test accuracy (\%) on analogy datasets. 
Results marked with * are not directly comparable, as they use a subset or the entire dataset to tune the model.
Results in bold represent the best accuracy excluding those marked with *. Underlined results show the best accuracy over all the models. Results in italics were taken from the original papers.}
\label{tab:analogy_result}
\end{table}

\begin{table*}[t]
\centering
\scalebox{0.8}{
\begin{tabular}{llcccccccccc}
\toprule
\multicolumn{2}{c}{\multirow{2}{*}{\textbf{Model}}} & \multicolumn{2}{c}{\textbf{BLESS}} & \multicolumn{2}{c}{\textbf{CogALexV}} & \multicolumn{2}{c}{\textbf{EVALution}} & \multicolumn{2}{c}{\textbf{K\&H+N}} & \multicolumn{2}{c}{\textbf{ROOT09}} \\
{} &{} &  macro &  micro &  macro &  micro &  macro &  micro &  macro &  micro &  macro &  micro \\
\midrule
\multirow{8}{*}{GloVe}  &            \textit{cat} &   92.9 &   93.3 &   42.8 &   73.5 &   56.9 &   58.3 &   88.8 &   94.9 &   86.3 &   86.5 \\
  &        \textit{cat+dot} &   \textbf{93.1} &   93.7 &   51.9 &   79.2 &   55.9 &   57.3 &   89.6 &   95.1 &   88.8 &   89.0 \\
  &   \textit{cat+dot+pair} &   91.8 &   92.6 &   56.4 &   81.1 &   58.1 &   59.6 &   89.4 &   95.7 &   89.2 &   89.4 \\
  &    \textit{cat+dot+rel} &   91.1 &   92.0 &   53.2 &   79.2 &   58.4 &   58.6 &   89.3 &   94.9 &   89.3 &   89.4 \\
  &           \textit{diff} &   91.0 &   91.5 &   39.2 &   70.8 &   55.6 &   56.9 &   87.0 &   94.4 &   85.9 &   86.3 \\
  &       \textit{diff+dot} &   92.3 &   92.9 &   50.6 &   78.5 &   56.5 &   57.9 &   88.3 &   94.8 &   88.6 &   88.9 \\
  &  \textit{diff+dot+pair} &   91.3 &   92.2 &   55.5 &   80.2 &   56.0 &   57.4 &   88.0 &   95.5 &   89.1 &   89.4 \\
  &   \textit{diff+dot+rel} &   91.1 &   91.8 &   52.8 &   78.6 &   56.9 &   57.9 &   87.4 &   94.6 &   87.7 &   88.1 \\
\midrule
\multirow{8}{*}{FastText}  &            \textit{cat} &   92.4 &   92.9 &   40.7 &   72.4 &   56.4 &   57.9 &   88.1 &   93.8 &   85.7 &   85.5 \\
  &        \textit{cat+dot} &   92.7 &   93.2 &   48.5 &   77.4 &   56.7 &   57.8 &   89.1 &   94.0 &   88.2 &   88.5 \\
  &   \textit{cat+dot+pair} &   90.9 &   91.5 &   53.0 &   79.3 &   56.1 &   58.2 &   88.3 &   94.3 &   87.7 &   87.8 \\
  &    \textit{cat+dot+rel} &   91.4 &   91.9 &   50.6 &   76.8 &   57.9 &   59.1 &   86.9 &   93.5 &   87.1 &   87.4 \\
  &           \textit{diff} &   90.7 &   91.2 &   39.7 &   70.2 &   53.2 &   55.5 &   85.8 &   93.3 &   85.5 &   86.0 \\
  &       \textit{diff+dot} &   92.3 &   92.9 &   49.1 &   77.8 &   55.2 &   57.4 &   86.5 &   93.6 &   88.5 &   88.9 \\
  &  \textit{diff+dot+pair} &   90.0 &   90.8 &   53.9 &   79.0 &   55.8 &   57.8 &   86.6 &   94.2 &   87.7 &   88.1 \\
  &   \textit{diff+dot+rel} &   90.6 &   91.3 &   53.6 &   78.2 &   57.1 &   58.0 &   86.3 &   93.4 &   86.9 &   87.4 \\
\midrule
\multirow{3}{*}{RelBERT}
  &                  Manual &   91.7 &   92.1 &   \textbf{71.2} &   \textbf{87.0} &   68.4 &   69.6 &   88.0 &   96.2 &   \textbf{90.9} &   \textbf{91.0} \\
  &              AutoPrompt &   91.9 &   92.4 &   68.5 &   85.1 &   \textbf{69.5} &   \textbf{70.5} &   \textbf{91.3} &   97.1 &   90.0 &   90.3 \\
  &                P-tuning &   91.3 &   91.8 &   67.8 &   84.9 &   69.1 &   70.2 &   88.5 &   96.3 &   89.8 &   89.9 \\
  \midrule
  \midrule
\multirow{2}{*}{SotA}  & 
                     LexNET &      - &   89.3 &  - &   - &   - &   60.0 &   - &   98.5 &   -  & 81.3 \\
  &              SphereRE &   - &   \textbf{93.8} &   - &   - &   - &  62.0 &   - &   \textbf{99.0} &   - &   86.1 \\
\bottomrule
\end{tabular}
}
\caption{Macro/micro F1 score (\%) for lexical relation classification.
}
\label{tab:classification_results}
\end{table*}

\subsection{Lexical Relation Classification}
\label{eval-relclass}

Table~\ref{tab:classification_results} summarizes the results of the lexical relation classification experiments, in terms of macro and micro averaged F1 score. The RelBERT models achieve the best results on all datasets except for BLESS and K\&H+N, where the performance of all models is rather close.
We observe a particularly large improvement over the word embedding and SotA models on the EVALution dataset. When comparing the different prompting strategies, we again find that the manual prompts perform surprisingly well, although the best results are now obtained with learned prompts in a few cases.

\begin{table}[t]
\centering
 \resizebox{\columnwidth}{!}{
\begin{tabular}{l@{\hspace{5pt}}c@{\hspace{5pt}}c@{\hspace{5pt}}c@{\hspace{5pt}}c@{\hspace{5pt}}c}
\toprule
{} & \textbf{BLESS} & \textbf{CogALex} & \textbf{EVAL} & \textbf{K\&H+N} & \textbf{ROOT09} \\
\midrule
rand  &   93.7\,(+0.3) &      94.3\,(-0.2) &                  - &   97.9\,(+0.2) &    91.2\,(-0.1) \\
mero  &   89.8\,(+1.4) &      72.9\,(+2.7) &       69.2\,(+1.9) &   74.5\,(+5.4) &               - \\
event &   86.5\,(-0.3) &                 - &                  - &              - &               - \\
hyp   &   94.1\,(+0.8) &      60.9\,(-0.7) &       61.7\,(-1.5) &   93.5\,(+5.0) &    83.0\,(-0.4) \\
cohyp &   96.4\,(+0.3) &                 - &                  - &   97.8\,(+1.2) &    97.4\,(-0.5) \\
attr  &   92.6\,(+0.3) &                 - &       84.7\,(+1.6) &              - &               - \\
poss  &              - &                 - &       67.1\,(-0.2) &              - &               - \\
ant   &              - &      76.8\,(-2.6) &       81.3\,(-0.9) &              - &               - \\
syn   &              - &      49.9\,(-0.6) &       53.6\,(+2.7) &              - &               - \\
\midrule
macro &   92.2\,(+0.5) &      71.0\,(-0.2) &       69.3\,(+0.9) &   90.9\,(+2.9) &    90.5\,(-0.4) \\
micro &   92.5\,(+0.4) &      86.9\,(-0.1) &       70.2\,(+0.6) &   97.2\,(+1.0) &    90.7\,(-0.3) \\
\bottomrule
\end{tabular}
}
\caption{Per-class F1 score of RelBERT trained without hypernymy relations and the absolute difference with respect to the original model (parentheses), along with the macro and micro averaged F1 for each dataset (\%).
}
\label{tab:hyp_result}
\end{table}

\begin{table}[t]
\centering
\footnotesize 
\scalebox{1.07}{
\begin{tabular}{l@{\hspace{15pt}}C{20pt}@{}C{20pt}@{\hspace{15pt}}C{20pt}@{}C{20pt}@{}C{20pt}}
\toprule
\multirow{2}{*}{\textbf{Model}} &  \multicolumn{2}{l}{\textbf{Google}} &  \multicolumn{3}{l}{\hspace{12pt}\textbf{BATS}}  \\
 &  \textbf{Mor} &  \textbf{Sem}&  \textbf{Mor} &  \textbf{Sem} &  \textbf{Lex}   \\
\midrule
FastText  &   95.4   &               98.1 &             90.4 &                 71.1 &               33.8 \\
\midrule
Manual      &   89.8   &95.8    &   87.0        &       66.2    &               75.1 \\
AutoPrompt  &   90.5   &85.1    &   85.3        &       59.8    &               68.0 \\
P-tuning    &   87.4   &78.1    &   82.9        &       60.9    &               61.8 \\
\bottomrule
\end{tabular}
}
\caption{Test accuracy for the high-level categories of BATS and Google, comparing FastText and RelBERT.}
\label{tab:breakdown}
\end{table}

\section{Analysis}

To better understand how relation embeddings are learned, in this section we analyze the model's performance in more detail.

\subsection{Training Data Overlap}
In our main experiments, RelBERT is trained using the SemEval 2012 Task 2 dataset. This dataset contains a broad range of semantic relations, including hypernymy and meronymy relations. This raises an important question: Does RelBERT provide us with a way to extract relational knowledge from the parameters of the pre-trained LM, or is it learning to construct relation embeddings from the triples in the training set? What is of particular interest is whether RelBERT is able to model types of relations that it has not seen during training. 
To answer this question, we conduct an additional experiment to evaluate RelBERT on lexical relation classification, using a version that was trained without the relations from the {\it Class Inclusion} category, which is the high-level category in the SemEval dataset that includes the {\it hypernymy} relation. Hypernymy is of particular interest, as it can be found across all the considered lexical relation classification datasets, which is itself a reflection of its central importance in lexical semantics.
In Table~\ref{tab:hyp_result}, we report the difference in performance compared to the original RelBERT model (i.e.\ the model that was fine-tuned on the full SemEval training set). 
As can be seen, the overall changes in performance are small, and the new version actually outperforms the original RelBERT model on a few datasets. In particular, hypernymy is still modelled well, confirming that RelBERT is able to generalize to unseen relations.

\begin{table*}[t]
\footnotesize
\begin{tabular}{@{}ll@{}}
\toprule
\textbf{Target} & \textbf{Nearest Neighbors}\\
\midrule
barista:coffee & baker:bread, brewer:beer, bartender:cocktail, winemaker:wine, bartender:drink, baker:cake\\
bag:plastic & bottle:plastic, bag:leather, container:plastic, box:plastic, jug:glass, bottle:glass\\
duck:duckling & chicken:chick, pig:piglet, cat:kitten, ox:calf, butterfly:larvae, bear:cub\\
cooked:raw & raw:cooked, regulated:unregulated, sober:drunk, loaded:unloaded, armed:unarmed, published:unpublished\\
chihuahua:dog & dachshund:dog, poodle:dog, terrier:dog, chinchilla:rodent, macaque:monkey, dalmatian:dog\\
dog:dogs & cat:cats, horse:horses, pig:pigs, rat:rats, wolf:wolves, monkey:monkeys\\
spy:espionage & pirate:piracy, robber:robbery, lobbyist:lobbying, scout:scouting, terrorist:terrorism, witch:witchcraft \\ 
\bottomrule
\end{tabular}
\caption{Nearest neighbors of selected word pairs, in terms of cosine similarity between RelBERT embeddings. Candidate word pairs are taken from the RELATIVE pair vocabulary.\label{tab:NNRelBERT}}
\end{table*}

As a further analysis, Table~\ref{tab:breakdown} shows a break-down of the Google and BATS analogy results, showing the average performance on each of the top-level categories from these datasets.\footnote{A full break-down showing the results for individual relations is provided in the appendix.} While RelBERT is outperformed by FastText on the morphological relations, it should be noted that the differences are small, while such relations are of a very different nature than those from the SemEval dataset. This confirms that RelBERT is able to model a broad range of relations, although it can be expected that better results would be possible by including task-specific training data into the fine-tuning process (e.g.\ including morphological relations for tasks where such relations matter).

\subsection{Language Model Comparison}

\begin{figure}[t]
    \centering
    \includegraphics[width=1.0\columnwidth]{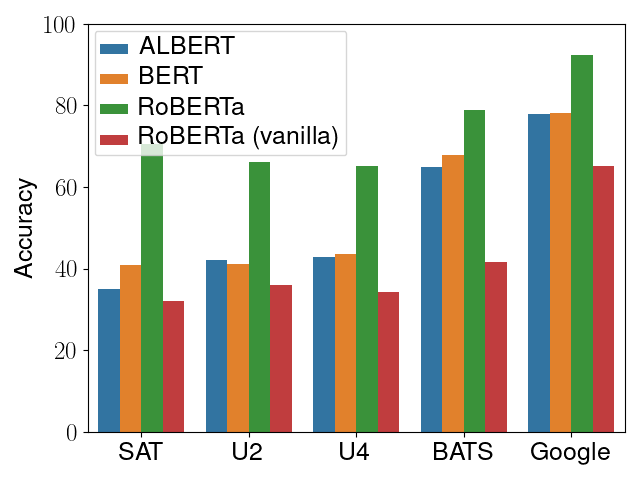}
    \caption{Test accuracy (\%) on analogy dataset of the vanilla RoBERTa model (i.e.\ without fine-tuning) and variants of RelBERT with different language models. Each variant uses the best manual prompt based on the SemEval tuning data. 
    }
    \label{fig:vanilla_custom}
\end{figure}

Figure~\ref{fig:vanilla_custom} compares the performance of RelBERT with that of the vanilla pre-trained RoBERTa model (i.e.\ when only the prompt is optimized). As can be seen, the fine-tuning process is critical for achieving good results.
In Figure~\ref{fig:vanilla_custom}, we also compare the performance of our main RelBERT model, which is based on RoBERTa, with versions that were instead initialized with  BERT \cite{devlin-etal-2019-bert} and ALBERT \cite{ALBERT}.\footnote{We used \texttt{bert-large-cased} and \texttt{albert-\allowbreak xlarge-\allowbreak v1} from the HuggingFace model hub. 
} RoBERTa clearly outperforms the other two LMs, which is in accordance with findings from the literature suggesting that RoBERTa captures more semantic knowledge \cite{li-etal-2020-linguistically,warstadt-etal-2020-learning}.

\subsection{Qualitative Analysis} To give further insight into the nature of RelBERT embeddings, Table \ref{tab:NNRelBERT} shows the nearest neighbors of some selected word pairs from the evaluation datasets. To this end, we computed RelBERT relation vectors for all pairs in the Wikipedia pre-trained RELATIVE vocabulary (over 1M pairs).\footnote{\url{https://github.com/pedrada88/relative}}
The neighbors are those word pairs whose RelBERT embedding has the highest cosine similarity within the full pair vocabulary. As can be seen, the neighbors mostly represent word pairs that are relationally similar, even for morphological relations (e.g.\ \emph{dog:dogs}), which are not present in the SemEval dataset.
A more extensive qualitative analysis, including a comparison with RELATIVE, is provided in the appendix.

\section{Conclusion}
We have proposed a strategy for learning relation embeddings, i.e.\ vector representations of pairs of words which capture their relationship. The main idea is to fine-tune a pre-trained language model using the relational similarity dataset from SemEval 2012 Task 2, which covers a broad range of semantic relations. In our experimental results, we found the resulting relation embeddings to be of high quality, outperforming state-of-the-art methods on several analogy and relation classification benchmarks. Among the models tested, we obtained the best results with RoBERTa, when using manually defined templates for encoding word pairs. Importantly, we found that high-quality relation embeddings can be obtained even for relations that are unlike those from the SemEval dataset, such as morphological and encyclopedic relations. This suggests that the knowledge captured by our relation embeddings is largely distilled from the pre-trained language model, rather than being acquired during training.

\section*{Acknowledgements}

Jose Camacho-Collados acknowledges support from the UKRI Future Leaders Fellowship scheme.

\bibliography{anthology,custom}
\bibliographystyle{acl_natbib}

\appendix

\section{Additional Experimental Results}
In this section, we show additional experimental results that complement the main results of the paper.

\subsection{Vanilla LM Comparison}
\begin{figure}[t]
    \centering
    \includegraphics[width=\columnwidth]{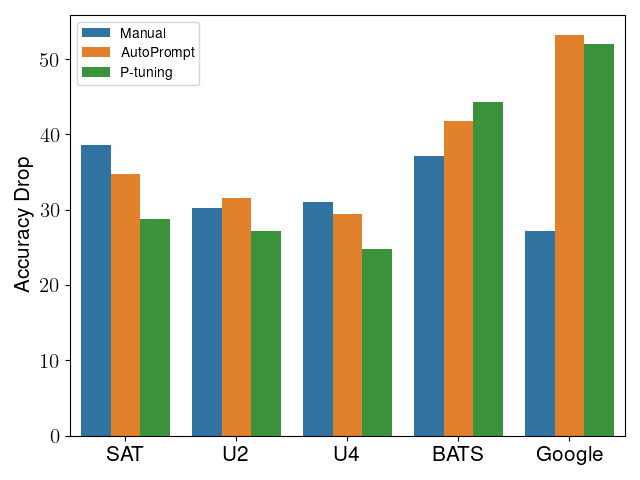}
    \caption{Test accuracy drop of the vanilla models without fine-tuning (measured in terms of absolute percentage points in comparison with RelBERT) on analogy datasets.}
    \label{fig:vanilla_sup_d}
\end{figure}

We show comparisons of versions of RelBERT with optimized prompt with/without finetuning.
Figure~\ref{fig:vanilla_sup_d} shows the absolute accuracy drop from RelBERT (i.e.\ the model with fine-tuning) to the vanilla RoBERTa model (i.e.\ without fine-tuning) with the same prompt. In all cases, the accuracy drop for the models without fine-tuning is substantial.

\subsection{Comparison with ALBERT \& BERT}
We use RoBERTa in our main experiments and here we train RelBERT with ALBERT and BERT instead, and evaluate them on both of the analogy and relation classification tasks.
Table~\ref{tab:analogy_sup} shows the accuracy on the analogy questions, while Table~\ref{tab:classification_sup} shows the accuracy on the relation classification task.
In both tasks, we can confirm that RoBERTa achieves the best performance within the LMs, by a relatively large margin in most cases.

\subsection{Word Embeddings}
Table~\ref{tab:analogy_result_sup} shows additional results of word embeddings on analogy test together with RelBERT results. We concatenate the RELATIVE and pair2vec vectors with the word vector difference. However, this does not lead to better results.

\section{Experimental Details and Model Configurations}
In this section, we explain models' configuration in the experiments, and details on RelBERT's training time.

\subsection{Prompting Configuration}

Table~\ref{fig:config} shows the best prompt configuration based on the validation loss for the SemEval 2012 Task 2 dataset in our main experiments using RoBERTa.

\begin{table}[t]
\centering
\scalebox{0.75}{
\begin{tabular}{lrrrrrr}
\toprule
\textbf{Model} &  \textbf{SAT\dag} &  \textbf{SAT} &  \textbf{U2} &  \textbf{U4} &  \textbf{Google} &  \textbf{BATS} \\
\midrule
\multicolumn{7}{l}{ALBERT} \\
$\cdot$ Manual     &              34.2 &          35.0 &         42.1 &         42.8 &             78.0 &           64.9 \\
$\cdot$ AutoPrompt &              34.0 &          35.0 &         36.4 &         33.8 &             25.0 &           30.5 \\
$\cdot$ P-tuning   &              32.4 &          32.6 &         33.8 &         33.6 &             35.0 &           37.8 \\ \midrule
\multicolumn{7}{l}{BERT} \\
$\cdot$ Manual     &              40.6 &          40.9 &         41.2 &         43.5 &             78.2 &           67.9 \\
$\cdot$ AutoPrompt &              36.4 &          36.5 &         36.8 &         35.4 &             51.6 &           43.5 \\
$\cdot$ P-tuning   &              38.0 &          38.0 &         38.2 &         37.0 &             56.6 &           45.3 \\ \midrule
\multicolumn{7}{l}{RoBERTa} \\
$\cdot$ Manual     &              \textbf{69.5} &          \textbf{70.6} &         \textbf{66.2} &         \textbf{65.3} &             \textbf{92.4} &           \textbf{78.8} \\
$\cdot$ AutoPrompt &              61.0 &          62.3 &         61.4 &         63.0 &             88.2 &           74.6 \\
$\cdot$ P-tuning   &              54.0 &          55.5 &         58.3 &         55.8 &             83.4 &           72.1 \\
\bottomrule
\end{tabular}
}
\caption{
Test accuracy (\%) of ALBERT, BERT, and RoBERTa on analogy datasets.
}
\label{tab:analogy_sup}
\end{table}

\begin{table*}[t]
\centering
\scalebox{0.78}{
\begin{tabular}{llcccccccccc}
\toprule
\multicolumn{2}{c}{\multirow{2}{*}{\textbf{Model}}} & \multicolumn{2}{c}{\textbf{BLESS}} & \multicolumn{2}{c}{\textbf{CogALexV}} & \multicolumn{2}{c}{\textbf{EVALution}} & \multicolumn{2}{c}{\textbf{K\&H+N}} & \multicolumn{2}{c}{\textbf{ROOT09}} \\
{} &{} &  macro &  micro &  macro &  micro &  macro &  micro &  macro &  micro &  macro &  micro \\
\midrule
\multirow{3}{*}{ALBERT}  &                  Manual &   86.2 &   87.1 &   54.9 &   81.1 &   62.6 &   62.2 &   82.6 &   91.7 &   86.4 &   86.8 \\
  &              AutoPrompt &   88.4 &   88.9 &   42.2 &   75.6 &   56.0 &   56.4 &   87.1 &   94.8 &   84.4 &   85.1 \\
  &                P-tuning &   90.1 &   90.6 &   44.9 &   73.1 &   58.2 &   59.7 &   90.2 &   95.9 &   85.9 &   85.9 \\
\midrule
\multirow{3}{*}{BERT}  &                  Manual &   90.9 &   91.2 &   65.2 &   83.4 &   67.8 &   68.3 &   91.6 &   97.6 &   90.1 &   90.4 \\
  &              AutoPrompt &   90.3 &   90.7 &   40.6 &   75.8 &   60.4 &   59.5 &   90.2 &   97.2 &   86.6 &   86.1 \\
  &                P-tuning &   87.6 &   88.0 &   52.7 &   79.2 &   61.9 &   63.3 &   86.2 &   95.1 &   85.2 &   85.3 \\
  \midrule
\multirow{3}{*}{RoBERTa}
  &                  Manual &   91.7 &   92.1 &   \textbf{71.2} &   \textbf{87.0} &   68.4 &   69.6 &   88.0 &   96.2 &   \textbf{90.9} &   \textbf{91.0} \\
  &              AutoPrompt &   \textbf{91.9} &   \textbf{92.4} &   68.5 &   85.1 &   \textbf{69.5} &   \textbf{70.5} &   \textbf{91.3} &   \textbf{97.1} &   90.0 &   90.3 \\
  &                P-tuning &   91.3 &   91.8 &   67.8 &   84.9 &   69.1 &   70.2 &   88.5 &   96.3 &   89.8 &   89.9 \\
\bottomrule
\end{tabular}
}
\caption{Macro/micro F1 score (\%) for lexical relation classification of ALBERT, BERT, and RoBERTa.
\label{tab:classification_sup}
}
\end{table*}

\begin{table*}[t]
\centering
\scalebox{0.83}{
\begin{tabular}{llcccccc}
\toprule
\multicolumn{2}{l}{\textbf{Model}}      & \textbf{SAT\dag} & \textbf{SAT}  & \textbf{U2}   & \textbf{U4}   & \textbf{Google} & \textbf{BATS} \\\midrule
\multirow{3}{*}{\rotatebox{90}{GloVe}}
                          & \textit{diff}        & 48.9              & 47.8          & 46.5          & 47.8          & {96.0}   & 68.7          \\
                          & \textit{diff}+\textit{rel}    & 45.9              & 40.4          & 46.9          & 35.4          & 87.6            & 67.3          \\
                          & \textit{diff}+\textit{pair}   & 35.1              & 33.8          & 29.4          & 30.6          & 78.0            & 56.3          \\\midrule
\multirow{3}{*}{\rotatebox{90}{FastText}}
                          & \textit{diff}        & 49.7              & 47.8          & 43.0          & 40.7          & \textbf{{96.6}}   & 72.0          \\
                          & \textit{diff}+\textit{rel}    & 37.3              & 35.9          & 39.5          & 35.6          & 85.8            & 67.5          \\
                          & \textit{diff}+\textit{pair}   & 33.4              & 33.5          & 27.2          & 28.7          & 75.4            & 52.1          \\\midrule
\multirow{3}{*}{\rotatebox{90}{RelBERT}}
& Manual     &              \textbf{{69.5}} &\textbf{{70.6}} & \textbf{{66.2}} &\textbf{{65.3}} &             92.4 &           \textbf{78.8} \\
& AutoPrompt &              61.0 &          62.3 &         61.4 &         63.0 &             88.2 &           74.6 \\
& P-tuning   &              54.0 &          55.5 &         58.3 &         55.8 &             83.4 &           72.1 \\\bottomrule
\end{tabular}
}
\caption{Test accuracy (\%) on analogy datasets (SAT\dag\: refers to the full SAT dataset).} 
\label{tab:analogy_result_sup}
\end{table*}

\begin{table}[h]
\centering
\scalebox{0.83}{
\begin{tabular}{llcccc}
\toprule
\textbf{Model}     &\textbf{Prompt}     & $\pi$ & $\tau$ & $\gamma$ & \textbf{template type} \\ \midrule
\multirow{3}{*}{BERT}
& Manual    & - & - & - & 3 \\
& AutoPrompt & 8     & 2      & 3 & -        \\
& P-tuning   & 8     & 2      & 2  & -        \\ \midrule
\multirow{3}{*}{ALBERT}
& Manual    & - & - & - & 4 \\
& AutoPrompt & 8     & 3      & 3  & -        \\
& P-tuning   & 8     & 2      & 3  & -        \\ \midrule
\multirow{3}{*}{RoBERTa}
& Manual    & - & - & - & 4 \\
& AutoPrompt & 9     & 2      & 2 & -         \\
& P-tuning   & 9     & 3      & 2 & -         \\ \bottomrule
\end{tabular}
}
\caption{Best prompting configuration.}
\label{fig:config}
\end{table}

\begin{table}[h]
\centering
\scalebox{0.83}{
\begin{tabular}{llccc}
\toprule
\textbf{Model}  
& \textbf{Data} & Manual    & AutoPrompt    & P-tuning \\\midrule
\multirow{5}{*}{ALBERT}
& BLESS &(1e-4, 150) &(1e-3, 200)     & (1e-4, 150)  \\
& CogA  &(1e-3, 100) &(1e-3, 100)     & (1e-3, 100)  \\
& EVAL  &(1e-4, 100) &(1e-3, 200)     & (1e-4, 100)  \\
& K\&H  &(1e-4, 150) &(1e-4, 150)     & (1e-4, 200)  \\
& ROOT  &(1e-5, 200) &(1e-4, 100)     & (1e-4, 150)  \\
\midrule
\multirow{5}{*}{BERT}
& BLESS &(1e-4, 200) &(1e-3, 100)     & (1e-3, 100)  \\
& CogA  &(1e-3, 100) &(1e-3, 100)     & (1e-3, 100)  \\
& EVAL  &(1e-5, 150) &(1e-3, 200)     & (1e-3, 100)  \\
& K\&H  &(1e-4, 200) &(1e-4, 200)     & (1e-3, 150)  \\
& ROOT  &(1e-5, 100) &(1e-3, 150)     & (1e-4, 150)  \\
\midrule
\multirow{5}{*}{RoBERTa}
& BLESS &(1e-5, 200) &(1e-3, 100)     & (1e-4, 200)  \\
& CogA  &(1e-3, 100) &(1e-3, 100)     & (1e-3, 100)  \\
& EVAL  &(1e-4, 150) &(1e-5, 100)     & (1e-5, 200)  \\
& K\&H  &(1e-3, 200) &(1e-3, 150)     & (1e-5, 200)  \\
& ROOT  &(1e-5, 100) &(1e-5, 200)     & (1e-3, 200)  \\
\bottomrule
\end{tabular}
}
\caption{
Best MLP configuration for the relation classification experiment. Each entry shows the learning rate and hidden layer size. Note that CogALex uses the default configuration due to the lack of validation set.
}
\label{fig:config_mlp}
\end{table}

\subsection{MLP Configuration in Relation Classification}
Table~\ref{fig:config_mlp} shows the best hyperparameters in the validation set of the MLPs for relation classification.

\subsection{Training Time}
Training a single RelBERT model with a custom prompt takes about half a day on two V100 GPUs. Additionally, to achieve prompt by AutoPrompt technique takes about a week on a single V100, while P-tuning takes 3 to 4 hours, also on a single V100.

\section{Implementation Details of AutoPrompt}
All the trigger tokens are initialized by mask tokens and updated based on the gradient of a loss function $L_t$. Concretely, let us denote the loss value with template $T$ as $L_t(T)$.
The candidate set for the $j$\textsuperscript{th} trigger is derived by 
\begin{align}
\tilde{\mathcal{W}_j} = \underset{w\in\mathcal{W}}{\text{top-}k} \big[ \bm{e}_w^T \nabla_j L_t(T) \big]
\end{align}
where the gradient is taken with respect to $j$\textsuperscript{th} trigger token and $\bm{e}_w$ is the input embedding for the word $w$. Then we evaluate each token based on the loss function as 
\begin{align}
    z_j = \underset{w\in \tilde{\mathcal{W}_j}}{\text{argmin}} \big[ L_t(\textit{rep}(T, j, w)) \big]
\end{align}
where $\textit{rep}(T, j, w)$ replaces the $j$\textsuperscript{th} token in $T$ by $w$ and $j$ is randomly chosen.
We ignore any candidates that do not improve current loss value to further enhance the prompt quality.

\section{Additional Analysis}
In this section, we analyze our experimental results based on prediction breakdown and provide an extended qualitative analysis.

\subsection{Prediction Breakdown}
Table~\ref{tab:bats_prediction} shows a detailed break-down of the BATS results.

\begin{table*}[t]
    \centering
\scalebox{0.83}{
\begin{tabular}{llrrrr}
\toprule
{} &   \textbf{Relation} &  \textbf{FastText} &  \textbf{Manual} &  \textbf{AutoPrompt} &  \textbf{P-tuning} \\
\midrule
\multirow{10}{*}{Encyclopedic}
&      UK city:county &               33.3 &             28.9 &                 28.9 &               40.0 \\
&      animal:shelter &               44.4 &             88.9 &                 77.8 &               84.4 \\
&        animal:sound &               80.0 &             86.7 &                 82.2 &               75.6 \\
&        animal:young &               53.3 &             62.2 &                 64.4 &               51.1 \\
&     country:capital &               82.2 &             37.8 &                 17.8 &               35.6 \\
&    country:language &               93.3 &             51.1 &                 55.6 &               51.1 \\
&         male:female &               88.9 &             60.0 &                 55.6 &               62.2 \\
&    name:nationality &               60.0 &             73.3 &                 51.1 &               40.0 \\
&     name:occupation &               86.7 &             75.6 &                 75.6 &               77.8 \\
&        things:color &               88.9 &             97.8 &                 88.9 &               91.1 \\
\midrule
\multirow{10}{*}{Lexical}    
&     antonyms:binary &               26.7 &             64.4 &                 68.9 &               77.8 \\
&   antonyms:gradable &               44.4 &             88.9 &                 93.3 &               88.9 \\
&   hypernyms:animals &               44.4 &             91.1 &                 80.0 &               55.6 \\
&      hypernyms:misc &               42.2 &             71.1 &                 60.0 &               64.4 \\
&       hyponyms:misc &               31.1 &             55.6 &                 55.6 &               48.9 \\
&     meronyms:member &               44.4 &             68.9 &                 48.9 &               53.3 \\
&       meronyms:part &               31.1 &             77.8 &                 71.1 &               55.6 \\
&  meronyms:substance &               26.7 &             75.6 &                 66.7 &               53.3 \\
&      synonyms:exact &               17.8 &             80.0 &                 71.1 &               66.7 \\
&  synonyms:intensity &               28.9 &             77.8 &                 64.4 &               53.3 \\
\midrule
\multirow{19}{*}{Morphological} 
&              adj+ly &               95.6 &             84.4 &                 88.9 &               82.2 \\
&            adj+ness &              100.0 &             97.8 &                 93.3 &               97.8 \\
&     adj:comparative &              100.0 &             97.8 &                100.0 &               91.1 \\
&     adj:superlative &               97.8 &            100.0 &                 93.3 &              100.0 \\
&           noun+less &               77.8 &            100.0 &                 97.8 &              100.0 \\
&            over+adj &               75.6 &             84.4 &                 80.0 &               82.2 \\
&              un+adj &               60.0 &             97.8 &                 91.1 &               97.8 \\
&      verb 3pSg:v+ed &              100.0 &             75.6 &                 84.4 &               68.9 \\
&       verb inf:3pSg &              100.0 &             93.3 &                 91.1 &               84.4 \\
&       verb inf:v+ed &              100.0 &             91.1 &                 91.1 &               88.9 \\
&      verb inf:v+ing &              100.0 &             97.8 &                 97.8 &               95.6 \\
&     verb v+ing:3pSg &               97.8 &             82.2 &                 68.9 &               68.9 \\
&     verb v+ing:v+ed &               97.8 &             86.7 &                 82.2 &               84.4 \\
&           verb+able &               97.8 &             93.3 &                 80.0 &               84.4 \\
&             verb+er &               95.6 &            100.0 &                 95.6 &               95.6 \\
&           verb+ment &               95.6 &             77.8 &                 77.8 &               62.2 \\
&           verb+tion &               84.4 &             77.8 &                 66.7 &               68.9 \\
&         noun:plural &               78.7 &             87.6 &                 88.8 &               69.7 \\
&             re+verb &               75.6 &             62.2 &                 86.7 &               66.7 \\
\bottomrule
\end{tabular}
}
\caption{Break-down of BATS results per relation type.}
\label{tab:bats_prediction}
\end{table*}

\subsection{Qualitative Analysis}
Tables \ref{tab:NNRelBERT_sup} shows the nearest neighbors of a number of selected word pairs, in terms of their RelBERT and RELATIVE embeddings. In both cases cosine similarity is used to compare the embeddings and the pair vocabulary of the RELATIVE model is used to determine the universe of candidate neighbors. 

The results for the RelBERT embeddings show their ability to capture a wide range of relations. 
In most cases the neighbors make sense, despite the fact that many of these relations are quite different from those in the SemEval dataset that was used for training RelBERT. 
The results for RELATIVE are in general much noisier, suggesting that RELATIVE embeddings fail to capture many types of relations. This is in particular the case for the morphological examples, although various issues can be observed for the other relations as well. 

\clearpage
\begin{sidewaysfigure*}

\scriptsize
\begin{tabular}{@{}l@{\hspace{4pt}}l@{\hspace{4pt}}ll@{}}
\toprule
\textbf{Category} &\textbf{Target} & \textbf{Nearest Neighbors RelBERT} & \textbf{Nearest Neighbors RELATIVE} \\
\midrule
\multirow{9}{*}{Commonsense} & barista:coffee & baker:bread, brewer:beer, bartender:cocktail, winemaker:wine, bartender:drink, baker:cake & venue:bar, restaurant:kitchen, restaurant:grill, nightclub:open, pub:bar, night:concert\\
&  restaurant:waitress& restaurant:waiter, diner:waitress, bar:bartender, hospital:nurse, courthouse:clerk, office:clerk & coincidentally:first, ironically:first, ironically:name, notably:three, however:new, instance:character\\
& car:garage & car:pit, plane:hangar, auto:garage, baby:crib, yacht:harbour, aircraft:hangar & shelter:house, elevator:building, worker:mine, worker:factory, plane:hangar, horse:stable\\
&  ice:melt & snow:melt, glacier:melt, ice:drift, crust:melt, polar ice:melt, ice:thaw & glacier:melt, snow:melt, water:freeze, crack:form, ice:surface, ice:freeze\\
& dolphin:swim & squid:swim, salmon:swim, shark:swim, fish:swim, horse:run, frog:leap & fisherman:fish, fisherman:catch, must:protect, diver:underwater, dog:human, scheme:make\\
&flower:fragrant & orchid:fragrant, cluster:fragrant, jewel:precious, jewel:valuable, soil:permeable, vegetation:abundant & flower:greenish, flower:white, 	flower:yellowish, flower:creamy, flower:pale yellow, flower:arrange\\
& coconut:milk & coconut:oil, goat:milk, grape:juice, palm:oil, olive:oil, camel:milk & dry:powder, mix:sugar, candy:chocolate, cook:fry, butter:oil, milk:coffee\\
&  bag:plastic & bottle:plastic, bag:leather, container:plastic, box:plastic, jug:glass, bottle:glass & tube:glass, bottle:plastic, typically:plastic, 	frame:steel, shoe:leather, wire:metal\\
& duck:duckling & chicken:chick, pig:piglet, cat:kitten, ox:calf, butterfly:larvae, bear:cub & adult:young, worker:queen, queen:worker, bird:fly, chick:adult, female:mat\\
\midrule
\multirow{1}{*}{Gender} & man:woman & men:women, male:female, father:mother, boy:girl, hero:heroine, king:queen & man:boy, woman:child, child:youth, officer:crew, bride:groom, child:teen\\ 
\midrule
\multirow{2}{*}{Antonymy} &  cooked:raw & raw:cooked, regulated:unregulated, sober:drunk, loaded:unloaded, armed:unarmed, published:unpublished & 	annual:biennial, raw:cook, herb:subshrub, aquatic:semi, shrub:small, fry:grill	\\
& normal:abnormal & ordinary:unusual, usual:unusual, acceptable:unacceptable, stable:unstable, rational:irrational, legal:illegal & acute:chronic, mouse:human, negative:positive, fat:muscle, cell:tissue, motor:sensory\\
\midrule
\multirow{4}{*}{Meronymy} & helicopter:rotor & helicopter:rotor blades, helicopter:wing, bicycle:wheel, motorcycle:wheel, airplane:engine, plane:engine & aircraft:engine, engine:crankshaft, landing gear:wheel, engine:camshaft, rotor:blade, aircraft:wing\\
& bat:wing & butterfly:wing, eagle:wing, angel:wing, cat:paw, lion:wings, fly:wing & mouse:tail, dog:like, dragon:like, human:robot, leopard:spot, cat:like\\
&  beer:alcohol & wine:alcohol, cider:alcohol, soda:sugar, beer:liquor, beer:malt, lager:alcoho & steel:carbon, cider:alcohol, humidity:average, rate:average, household:non, consume:beer\\
& oxygen:atmosphere & helium:atmosphere, hydrogen:atmosphere, nitrogen:atmosphere, methane:atmosphere, carbon:atmosphere 
& carbon dioxide:atmosphere, cloud:atmosphere, methane:atmosphere, nitrogen:soil, gas:atmosphere\\ 
\midrule
\multirow{3}{*}{Hypernymy}  & chihuahua:dog & dachshund:dog, poodle:dog, terrier:dog, chinchilla:rodent, macaque:monkey, dalmatian:dog & julie:katy, tench:pike, catfish:pike, sunfish:perch, salmonid:salmon, 	raw:marinate\\
& pelican:bird & toucan:bird, puffin:bird, egret:bird, peacock:bird, grouse:bird, pigeon:bird & drinking:contaminate, drinking:source, 	pelican:distinctive, boiling:pour, aquifer:table, fresh:source\\
& tennis:sport & hockey:sport, soccer:sport, volleyball:sport, cricket:sport, golf:sport, football:sport	 & hockey:sport, golf:sport, badminton:sport, boxing:sport, rowing:sport, volleyball:sport\\
\midrule
\multirow{3}{*}{Morphology} & dog:dogs & cat:cats, horse:horses, pig:pigs, rat:rats, wolf:wolves, monkey:monkeys & shepherd:dog, landrace:breed, like:dog, farm:breed, breed:animal, captive:release\\
& tall:tallest & strong:strongest, short:shortest, smart:smartest, weak:weakest, big:biggest, small:smallest & rank:world, 	summit:nato, redistricting:district, delegation:congress, debate:congress \\ 
& spy:espionage & pirate:piracy, robber:robbery, lobbyist:lobbying, scout:scouting, terrorist:terrorism, witch:witchcraft & group:call, crime:criminal, action:involve, cop:police, action:one, group:make\\
\bottomrule
\end{tabular}
\caption{Nearest neighbors of selected word pairs, in terms of cosine similarity between RelBERT embeddings. Candidate word pairs are taken from the RELATIVE pair vocabulary.\label{tab:NNRelBERT_sup}}

\end{sidewaysfigure*}
\end{document}